\documentclass[conference]{IEEEtran}
\IEEEoverridecommandlockouts
\usepackage{cite}
\usepackage{amsmath,amssymb,amsfonts}
\usepackage{algorithmic}
\usepackage{graphicx}
\usepackage{textcomp}
\usepackage{xcolor}
\usepackage{booktabs}
\usepackage{caption}
\usepackage{subcaption}
\usepackage{color}    

\usepackage[hidelinks]{hyperref}
\usepackage{url}
\def\method{CNODE}

\def\BibTeX{{\rm B\kern-.05em{\sc i\kern-.025em b}\kern-.08em
    T\kern-.1667em\lower.7ex\hbox{E}\kern-.125emX}}
\begin{document}

\title{Conditional Neural ODE for Longitudinal Parkinson’s Disease Progression Forecasting\\
}


\author{%
\parbox{0.96\linewidth}{\centering
Xiaoda Wang$^{\ast}$, Yuji Zhao$^{\ddagger}$, Kaiqiao Han$^{\dagger}$, Xiao Luo$^{\|}$,
Sanne van Rooij$^{\ast}$, Jennifer Stevens$^{\ast}$,\\
Lifang He$^{\S}$, Liang Zhan$^{\P}$,
Yizhou Sun$^{\dagger}$, Wei Wang$^{\dagger}$, Carl Yang$^{\ast}$\\[2pt]
{\small
$^{\ast}$\textit{Emory University}\quad
$^{\dagger}$\textit{University of California, Los Angeles}\quad
$^{\ddagger}$\textit{Illinois Institute of Technology}\\
$^{\S}$\textit{Lehigh University}\quad
$^{\P}$\textit{University of Pittsburgh}\quad
$^{\|}$\textit{University of Wisconsin--Madison}
}\\[4pt]
\small
\nolinkurl{{xiaoda.wang,sanne.van.rooij,jennifer.stevens,j.carlyang}@emory.edu}\quad
\nolinkurl{yzhao104@hawk.iit.edu}\\
\nolinkurl{{kqhan,yzsun,weiwang}@cs.ucla.edu}\quad
\nolinkurl{xiao.luo@wisc.edu}\quad
\nolinkurl{lih319@lehigh.edu}\quad
\nolinkurl{liang.zhan@pitt.edu}
}%
}

\maketitle

\begin{abstract}
Parkinson’s disease (PD) shows heterogeneous, evolving brain-morphometry patterns. Modeling these longitudinal trajectories enables mechanistic insight, treatment development, and individualized ‘digital-twin’ forecasting.
However, existing methods usually adopt recurrent neural networks and transformer architectures, which rely on discrete, regularly sampled data while struggling to handle irregular and sparse magnetic resonance imaging (MRI) in PD cohorts. Moreover, these methods have difficulty capturing individual heterogeneity including variations in disease onset, progression rate, and symptom severity, which is a hallmark of PD. 
To address these challenges, we propose CNODE (\underline{C}onditional \underline{N}eural \underline{ODE}), a novel framework for continuous, individualized PD progression forecasting. The core of CNODE is to model morphological brain changes as continuous temporal processes using a neural ODE model. In addition, we jointly learn patient-specific initial time and progress speed to align individual trajectories into a shared progression trajectory. 
We validate CNODE on the Parkinson’s Progression Markers Initiative (PPMI) dataset. Experimental results show that our method outperforms state-of-the-art baselines in forecasting longitudinal PD progression.

\end{abstract}


\section{Introduction}
%
PD is a common neurodegenerative disorder with heterogeneous motor and non-motor manifestations and substantial societal burden~\cite{armstrong2020diagnosis,aarsland2021parkinson}.
Clinically, PD presents with a wide spectrum of motor symptoms and non-motor symptoms~\cite{gratwicke2015parkinson,sousani2024insights}. 
The progression of PD along with these symptoms varies across individuals, which reflects heterogeneous pathophysiological changes over years~\cite{wielinski2005falls, xie2025kerap, xie2025hypkg}.
And magnetic resonance imaging (MRI) has become a valuable tool to study morphological alterations associated with PD~\cite{herold2017functional,zhao2024morph}. Recent advances~\cite{han2021inflammation} in structural and functional MRI analyses have revealed PD’s neuroanatomical correlates, including changes in deep brain structures, and shifts in functional connectivity.

However, capturing the full PD progression is challenging. In particular, we expect effective analytical methods that can model continuous, and highly individualized trajectories of brain morphology, which remain difficult due to the nature of PD.
First, patients undergo MRI at varying intervals in real-world clinical settings. Traditional deep learning methods such as recurrent neural networks (RNNs) and transformer-based architectures typically assume regularly sampled data, making them poorly suited for handling inherently continuous data sampled with varying intervals or missing values.
Second, the disease onset age, progression rate, and symptom severity vary greatly among patients, which indicates distinct clinical trajectories even in relatively homogeneous cohorts~\cite{lian2024personalized}. 
When studying longitudinal progression, it is highly anticipated to capture the intrinsic heterogeneity of PD progression.


To address these challenges, we propose \method{} (\underline{C}onditional \underline{N}eural \underline{ODE}), a novel framework designed to forecast longitudinal PD progression from irregular and sparsely sampled MRI scans. Specifically, \method{} models disease evolution as a continuous trajectory in a shared latent space, while jointly learning patient-specific initial time and progression speed as conditioning variables. Then \method{} captures individualized disease dynamics and effectively aligns all patients onto a unified progression trajectory, enabling robust prediction and analysis of heterogeneous disease courses. 

In summary, our main contributions are as follows: 
\begin{itemize}
    \item We are the first to recognize the importance of modeling the continuous progression for PD and propose \method{} to model continuous PD progression, effectively addressing the challenges of irregular sampled MRI data.
    \item We harness subcortical segmentations from T1-weighted MRI to extract vertex-specific shape features. Instead of modeling the subject-wise whole brain shape directly, \method{} extracts the vertex-wise medial thickness.
    \item \method{} aligns all patient into a shared progression trajectory by jointly learning patient-specific initial time and progression speed. \method{} also leverages contrastive learning for trajectory alignment.
    \item We validate \method{} on the PPMI dataset, showing that our method outperforms state-of-the-art baselines and provides insights into personalized disease trajectories forecasting.
\end{itemize}

\begin{figure*}[t]
    \centering
    \includegraphics[width=0.85\linewidth]{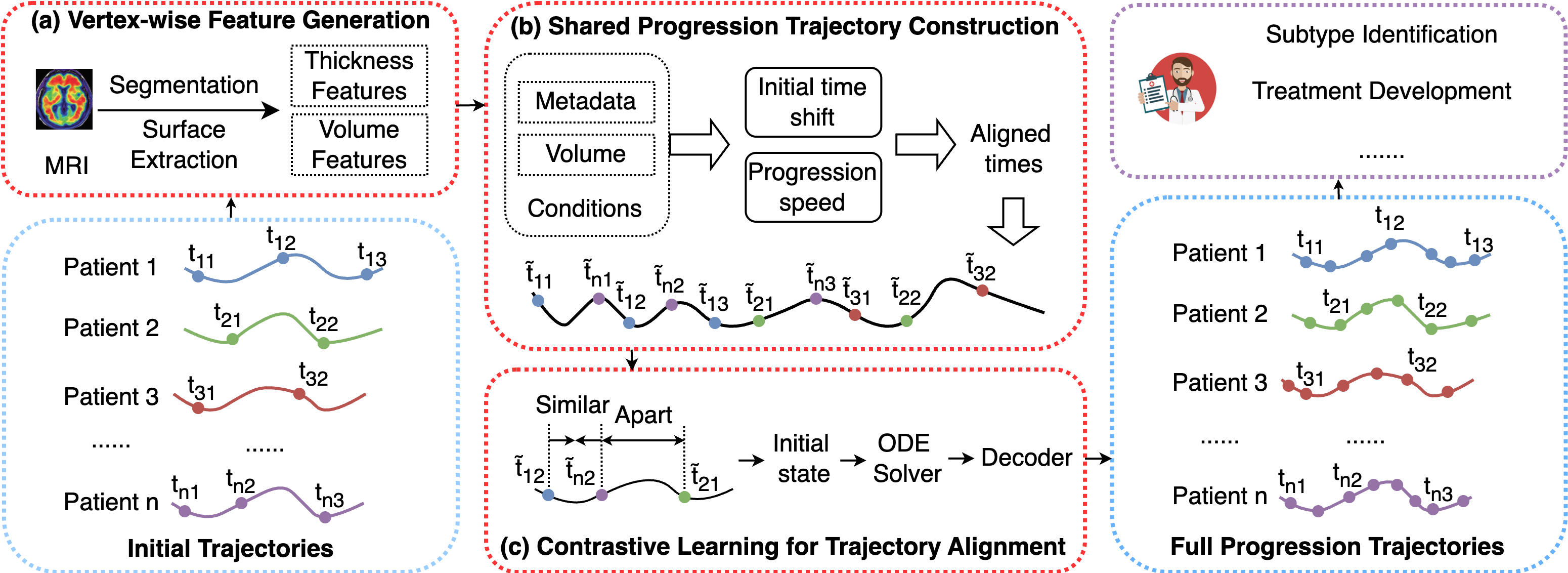}
    \caption{Overview of \method{}. (a) Firstly, we extract vertex-wise \textit{medial thickness} from MRI, with subcortical structure volumes. (b) Secondly, we construct a shared progression trajectory. (c) Then we employ contrastive learning to align time distributions. Finally, we model the continuous evolution with Neural ODE.}
    \label{fig:overview}
\end{figure*}


\section{Methodology}
In this section, we present \method{}, a Conditional Neural ODE-based framework to model the longitudinal PD progression as shown in Figure~\ref{fig:overview}. 
First, we extract vertex-wise \textit{medial thickness} from MRI scans, with subcortical structure volumes and metadata as condition features. Second, to address sparse visit times, we construct a shared progression trajectory. We also employ contrastive learning, leveraging shape feature similarity to enforce aligned time distributions. Then we model the continuous evolution of shape features with Neural ODE~\cite{liu2025graph}.

\noindent \textbf{Problem Definition.}
We focus on the problem of continuous and longitudinal PD progression forecasting.
Let us consider a set of $N$ subjects (patients) denoted by $S = \{ s_i \}_{i=1}^{N}$.
Each subject $s_i$ is has $K_i$ observed timestamps
\(
\mathcal{T}_i = \{ t_{i,1}, t_{i,2}, \ldots, t_{i,K_i} \},
\)
where these timestamps may be non-uniformly spaced and can take arbitrary continuous values. 
At each time $t_{i,k}$, we extract \textit{shape features} $X_{i,k} \in \mathbb{R}^D$, where $D$ is the dimensionality of the feature vector.
Additionally, each subject $s_i$ has associated \textit{volume features} $C_i$ and \textit{metadata} $M_i$.
Given a limited number of observed shape features $\{X_{i,0:n_i}\}$ for each subject $s_i$, along with the subject-specific volume features $C_i$ and metadata $M_i$,
our goal is to forecast the future shape features $\{X_{i,n_i+1 : T_i}\}$ at arbitrary future times.
Formally, we aim to learn a model
\(
\hat{X}_{i,n_i+1:T_i} = f\bigl(\{X_{i,0:n_i}\}, C_i, M_i; \theta \bigr),
\)
where $\theta$ represents the model parameters.

\vspace{-0.7em}

\subsection{Vertex-wise Feature Generation}
Figure~\ref{fig:overview} illustrates our subcortical mesh-based feature generation process. 
We extract morphometrical features with respect to subcortical and cortical brain structures.
We focus on \emph{medial thickness} as the primary shape feature. Cortical medial thickness is extracted at each vertex, and the average thickness is computed across all vertices within each of the 34 predefined bilateral cortical per hemisphere, as defined by the Desikan-Killiany atlas. The resulting vertex-wise thickness values at time \( t_{i,k} \) form the shape feature vector \( X_{i,k} \) for subject \( s_i \).
We also extract \emph{subcortical structure volumes}. Given that some subjects have only one or two visits, the volumes features $C_{i}$ and metadata $M_i$ (age, gender) from the first visit serve as the condition features .

\subsection{Shared Progression Trajectory Construction}
Longitudinal visit times are sparse and irregular, which impedes learning a reliable disease timeline. We leverage subcortical volumes and metadata—informative for PD staging~\cite{zhao2022learning,gu2021changes}—to \emph{align} all subjects onto a shared progression axis by estimating (i) an onset shift and (ii) an individual speed. Two lightweight regressors $f_{\theta_\tau}$ and $f_{\theta_\gamma}$ map patient metadata $M_i$ and condition features $C_i$ to these alignment parameters:
\begin{equation}
\tau_i = f_{\theta_\tau}(M_i, C_i), \qquad
\gamma_i = f_{\theta_\gamma}(M_i, C_i).
\end{equation}
Given a raw visit time $t_{i,k}$ and the subject's first visit $t_{i,1}$, the aligned time is
\begin{equation}
\tilde{t}_{i,k} \;=\; \tau_i \;+\; \gamma_i \,\bigl(t_{i,k}-t_{i,1}\bigr),
\end{equation}
which standardizes trajectories by correcting onset heterogeneity and scaling progression rates.

\begin{table*}[h]
\setlength{\tabcolsep}{6pt} 
\centering
\caption{\textbf{Comparison of PD progression forecasting performance.} All metrics are presented as averages $\pm$ standard deviations under 5-fold cross-validation. The best performances are in bold.}
\label{tab:pd_forecasting_results}
\begin{tabular}{lccc}
\toprule
\textbf{Model} & \textbf{MSE} $\downarrow$ & \textbf{RMSE} $\downarrow$ & \textbf{$R^2$} $\uparrow$ \\
\midrule
GRU & $0.0281 \pm 0.0012$ &
       $0.1675 \pm 0.0036$ &
       $0.8112 \pm 0.0082$ \\

RNN & $0.0305 \pm 0.0027$ {\tiny\color{teal}$\uparrow 0.0024$} & 
      $0.1745 \pm 0.0076$ {\tiny\color{teal}$\uparrow 0.0070$} &
      $0.7947 \pm 0.0179$ {\tiny\color{red}$\downarrow 0.0165$} \\

LSTM & $0.0292 \pm 0.0019$ {\tiny\color{teal}$\uparrow 0.0011$} &
       $0.1708 \pm 0.0056$ {\tiny\color{teal}$\uparrow 0.0033$} &
       $0.8037 \pm 0.0130$ {\tiny\color{red}$\downarrow 0.0075$} \\

Transformer & $0.0291 \pm 0.0010$ {\tiny\color{teal}$\uparrow 0.0010$} &
              $0.1705 \pm 0.0030$ {\tiny\color{teal}$\uparrow 0.0030$} &
              $0.8038 \pm 0.0069$ {\tiny\color{red}$\downarrow 0.0074$} \\

Neural ODE & $0.0283 \pm 0.0019$ {\tiny\color{teal}$\uparrow 0.0002$} &
             $0.1632 \pm 0.0043$ {\tiny\color{red}$\downarrow 0.0043$} &
             $0.7907 \pm 0.0147$ {\tiny\color{red}$\downarrow 0.0205$} \\

LLMTime & $0.0282 \pm 0.0105$ {\tiny\color{teal}$\uparrow 0.0001$} &
          $0.1652 \pm 0.0226$ {\tiny\color{red}$\downarrow 0.0023$} &
          $0.8041 \pm 0.0739$ {\tiny\color{red}$\downarrow 0.0071$} \\

\midrule


\method{} (Ours) & $\mathbf{0.0258 \pm 0.0001}$ {\tiny\color{red}$\downarrow 0.0023$} &
                   $\mathbf{0.1606 \pm 0.0003}$ {\tiny\color{red}$\downarrow 0.0069$} &
                   $\mathbf{0.8260 \pm 0.0006}$ {\tiny\color{teal}$\uparrow 0.0148$} \\
\bottomrule
\end{tabular}
\end{table*}

\subsection{Contrastive Learning for Shared Trajectory Alignment}
Aligned times lack ground truth. To regularize them, we impose a metadata-aware contrastive objective so that visits with \emph{similar shape features} are temporally close, and dissimilar ones are far. We declare a pair $(i,k)$ and $(j,\ell)$ as \emph{positive} if the cosine similarity between their shape features $X_{i,k}$ and $X_{j,\ell}$ exceeds a threshold $\delta$; otherwise it is \emph{negative}. We weight each pair by the similarity of their conditions:
\begin{equation}\label{eq:weight}
w_{ij} \;=\; \mathrm{sim}\!\bigl((M_i,C_i),(M_j,C_j)\bigr).
\end{equation}
Let $\mathrm{s}(\tilde{t}_i,\tilde{t}_j)$ denote cosine similarity between aligned times. Using temperature $\tau$, the metadata-weighted InfoNCE loss is
\begin{equation}\label{eq:infoNCE}
\mathcal{L}_{\mathrm{contrast}}
=
- \sum_{i}
\sum_{\substack{j:\ \text{positive to } i}}
w_{ij}
\;\log 
\frac{
\exp\!\bigl(\mathrm{s}(\tilde{t}_i,\tilde{t}_j)/\tau\bigr)
}{
\sum_{k}\exp\!\bigl(\mathrm{s}(\tilde{t}_i,\tilde{t}_k)/\tau\bigr)
}.
\end{equation}
This encourages anchors to be close to positive visits while contrasting against all others, with stronger influence from clinically similar patients via $w_{ij}$. The result is a shared progression timeline that reflects both morphology and patient condition under sparse, irregular sampling.

\subsection{Neural ODE for Longitudinal Progression Modeling}
Given the aligned visit times $\tilde{t}_{i,k}$, we model subject-specific continuous trajectories in a latent space and decode them back to shape features.

\subsubsection{Initial State Encoder}
For subject $i$, the encoder summarizes the observed trajectory into a Gaussian posterior over the initial latent state:
\begin{equation}
q_\phi\!\left(z_{i,0}\mid X_{i,0:n_i}\right)=\mathcal{N}\!\big(\mu_i,\operatorname{diag}(\sigma_i^2)\big)
\end{equation}
\begin{equation}
\quad (\mu_i,\log\sigma_i)=\mathrm{Enc}_\phi\!\left(X_{i,0:n_i}\right).
\end{equation}

\subsubsection{ODE Generative Model}
Latent dynamics evolve continuously along the aligned timeline $\tilde{\mathbf t}_i=\{\tilde{t}_{i,1},\dots,\tilde{t}_{i,n_i}\}$:
\begin{equation}
z_i(\tilde{\mathbf t}_i)\;=\;\texttt{ODESolve}\!\big(g_\vartheta,\; z_{i,0},\; \tilde{\mathbf t}_i\big),
\end{equation}
where $g_\vartheta$ is a neural ODE field parameterizing disease progression.

\subsubsection{Decoder}
At each aligned time, the decoder reconstructs shape features:
\begin{equation}
\hat{X}_{i,k}=D_\theta\!\big(z_i(\tilde{t}_{i,k})\big)
\end{equation}
\begin{equation}
p\!\left(X_{i,k}\mid z_i(\tilde{t}_{i,k})\right)=\mathcal{N}\!\big(\hat{X}_{i,k},\sigma_x^2 I\big).
\end{equation}
This yields flexible, continuous-time forecasts under sparse and irregular sampling.

\subsection{Training}
We jointly optimize the encoder, ODE field, and decoder end-to-end with reconstruction, regularization, and alignment terms. Over all subjects and observed visits:
\begin{equation}
\mathcal{L}_{\text{MSE}}=\frac{1}{\sum_i n_i}\sum_{i}\sum_{k=1}^{n_i}\big\|X_{i,k}-\hat{X}_{i,k}\big\|_2^2,
\end{equation}
\begin{equation}
\mathcal{L}_{\text{KL}}=\sum_i D_{\mathrm{KL}}\!\left(\mathcal{N}(\mu_i,\operatorname{diag}(\sigma_i^2))\,\big\|\,\mathcal{N}(0,I)\right),
\end{equation}
The final objective is
\begin{equation}
\mathcal{L} \;=\; \mathcal{L}_{\text{MSE}} \;+\; \lambda_{\text{KL}}\,\mathcal{L}_{\text{KL}} \;+\; \lambda_{\text{C}}\,\mathcal{L}_{\text{contrast}},
\end{equation}
with $\lambda_{\text{KL}}$ and $\lambda_{\text{C}}$ balancing regularization and alignment.

\section{Experiment}

In this section, we comprehensively evaluate the effectiveness of the proposed \method{} framework through extensive experiments on longitudinal Parkinson’s disease (PD) data.

\noindent 

\subsection{Dataset Description and Implementation Details}
\subsubsection{Dataset Description} We evaluate \method{} using the PPMI dataset~\cite{marek2011parkinson}, a comprehensive, multi-center study aimed at identifying progression markers for Parkinson’s disease. For this study, we utilize T1-weighted MRI, which provides high-resolution anatomical details of the brain. The vertex-wise features were extracted from 68 sub-cortical brain regions using FreeSurfer. The final cohort includes 161 PD subjects, with 50 individuals having three visits and another 111 having two visits. The average visit interval is 1.11 years, with a maximum of 2.27 years and a minimum of 0.61 years.

\subsubsection{Evaluation Metrics}
The effectiveness of each model was evaluated using Mean Squared Error (MSE), Root Mean Squared Error (RMSE), and the Coefficient of Determination ($R^2$). MSE quantifies prediction error, RMSE provides magnitude, and $R^2$ measures the variance explained. Lower MSE and RMSE indicate better accuracy, while a higher $R^2$ suggests improved trend capture.

\subsection{Overall PD Progression Forecasting Performance}

We evaluated our model's performance in forecasting PD progression against several baseline models:

\begin{itemize}
    \item \textbf{Recurrent Neural Network (RNN)}~\cite{elman1990finding}: A classical sequential model capturing temporal dependencies.
    \item \textbf{Long Short-Term Memory (LSTM)}~\cite{hochreiter1997long} and \textbf{Gated Recurrent Unit (GRU)}~\cite{cho2014learning}: Popular RNN variants designed to address long-term dependency and vanishing gradient issues.
    \item \textbf{Transformer}~\cite{vaswani2017attention}: A self-attention-based model that can flexibly model long-range dependencies in temporal sequences.
    \item \textbf{Neural Ordinary Differential Equation (Neural ODE)}~\cite{chen2018neural}: A continuous-time latent dynamics model for irregularly sampled time series.
    \item \textbf{LLMTime}~\cite{gruver2023large}: Uses large language models to perform time series forecasting in a zero-shot manner.

\end{itemize}

For baselines designed for regularly spaced intervals, we set the interval to one year.
%
Table~\ref{tab:pd_forecasting_results} summarizes the forecasting results for all models. As shown, \method{} consistently achieves superior performance across all evaluation metrics, including the lowest Mean Squared Error (MSE), lowest Root Mean Squared Error (RMSE), and highest coefficient of determination ($R^2$). These results highlight the effectiveness of our approach in capturing the continuous and heterogeneous progression of PD, even under irregular and sparse sampling conditions.

\subsection{Hyperparameter Analysis}

\begin{figure}[h]
    \centering
    \includegraphics[width=0.8\linewidth]{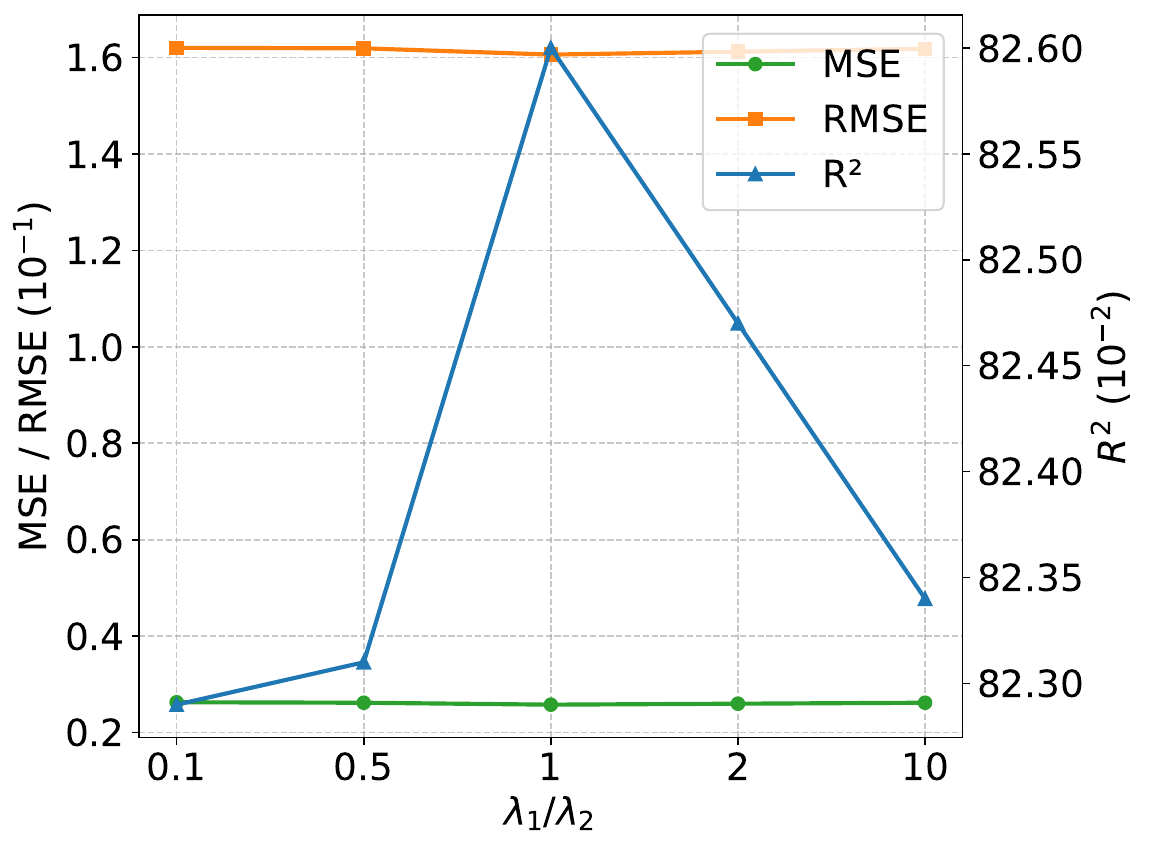}
    \caption{Effect of $\lambda_1/\lambda_2$ on CNODE performance.}
    \label{fig:hyperpara}
\end{figure}

We performed a hyperparameter analysis to examine how the balance between the KL divergence loss ($\lambda_1$) and the contrastive loss ($\lambda_2$) affects model performance. As shown in Figure~\ref{fig:hyperpara}, we varied the ratio $\lambda_1/\lambda_2$ across a broad range. The results demonstrate that CNODE achieves the best forecasting accuracy—reflected by lower MSE/RMSE and higher $R^2$—when $\lambda_1/\lambda_2$ is set to 1:1. These findings highlight the importance of properly weighting the KL and contrastive objectives for learning effective latent representations and robust trajectory alignment.

\section{Conclusion}
In this work, we introduced \method{}, a Conditional Neural ODE framework that models continuous Parkinson’s disease progression. By harnessing subcortical shape features and aligning patients within a shared progression trajectory, \method{} captures individualized disease dynamics while addressing common challenges such as missing timepoints and heterogeneous progression rates. 
Experimental results demonstrate the superiority of our proposed \method{} over state-of-the-art methods, highlighting its potential for personalized disease monitoring and therapeutic decision support.

\section{Acknowledgement}

This research was partially supported by the U.S. National Science Foundation under Award Numbers 2319449, 2312502, 2442172, 2045848, 2215789, 2319450, 2319451, 2106859, 2200274, 2312501, 2211557, 2119643, 2303037 and 2531008; by the U.S. National Institutes of Health under Award Numbers U54HG012517, U24DK097771, U54OD036472, and OT2OD038003; and by the US National Institute of Diabetes and Digestive and Kidney Diseases of the US National Institutes of Health under Award Number K25DK135913. 
Additional support was provided by SRC JUMP 2.0 Center, Amazon, NEC, Optum AI, Amazon Research Awards, and Snapchat Gifts.

\bibliography{reference}
\bibliographystyle{splncs04}

\end{document}